\documentclass{article}
\usepackage{ijcai19}

\usepackage{times}
\usepackage{soul}
\usepackage{url}
\usepackage[hidelinks]{hyperref}
\usepackage[utf8]{inputenc}
\usepackage[small]{caption}
\usepackage{graphicx}
\usepackage{amsmath}
\usepackage{booktabs}
\usepackage{algorithm}
\usepackage{algorithmic}
\title{Deep Active Learning for Anchor User Prediction}

\author{
Anfeng Cheng$^{1,2}$
\and
Chuan Zhou$^{1,2}$\footnote{Contact Author}\and
Hong Yang$^3$\and
Jia Wu$^4$\and
Lei Li$^5$\and
\\Jianlong Tan$^{1,2}$\And
Li Guo$^{1,2}$
\affiliations
$^1$Institute of Information Engineering, Chinese Academy of Sciences, Beijing, China\\
$^2$School of Cyber Security, University of Chinese Academy of Sciences, Beijing, China\\
$^3$Centre for Artificial Intelligence, School of Software, FEIT, University of Technology Sydney\\
$^4$Department of Computing, Macquarie University, Sydney, Australia\\
$^5$School of Computer Science and Information Engineering, Hefei University of Technology, China
\emails
\{chenganfeng, zhouchuan, tanjianlong, guoli\}@iie.ac.cn,
hong.yang@student.uts.edu.au,
jia.wu@mq.edu.au,
lilei@hfut.edu.cn
}

\usepackage{booktabs}
\usepackage{amsmath}
\usepackage{bm}
\usepackage{amsfonts}
\usepackage{multirow}
\usepackage{subfigure}
\usepackage{algorithm}
\usepackage{algorithmic}
\usepackage{array}
\usepackage{flushend}
\usepackage{color}

\begin{document}

\maketitle

\begin{abstract}
  Predicting pairs of anchor users plays an important role in the cross-network analysis. Due to the expensive costs of labeling anchor users for training prediction models, we consider in this paper the problem of minimizing the number of user pairs across multiple networks for labeling as to improve the accuracy of the prediction. To this end, we present a deep active learning model for anchor user prediction (DAL\begin{small}AUP\end{small} for short). However, active learning for anchor user sampling meets the challenges of non-i.i.d. user pair data caused by network structures and the correlation among anchor or non-anchor user pairs. To solve the challenges, DAL\begin{small}AUP\end{small} uses a couple of neural networks with shared-parameter to obtain the vector representations of user pairs, and ensembles three query strategies to select the most informative user pairs for labeling and model training. Experiments on real-world social network data demonstrate that DAL\begin{small}AUP\end{small} outperforms the state-of-the-art approaches.
\end{abstract}

\section{Introduction}
Social network users often join multiple social networks to obtain versatile services. For example, a user can be simultaneously active in both \textit{Foursquare} and \textit{Twitter}. These users are often termed as \textit{anchor users} who often generate rich data for cross-network analysis. Advanced services for anchor users include cross-network recommendation \cite{li2014matching}, link prediction \cite{zhang2013predicting} and information diffusion analysis \cite{peng2013predicting}. However, cross-network data generated from the anchor users are difficult to collect, because these users rarely use the same identities in different social networks, which poses difficulties for advancing cross network applications. Therefore, identifying anchor users across multiple social networks has attracted increasing research interests in recent years~\cite{man2016predict}.

To identify anchor users, previous work~\cite{liu2013s,riederer2016linking} try to collect and combine users' demographic data and daily-generated content data for estimation. For example, they often combine users' registration profile data such as names, genders and locations, and their daily-generated tweets, posts, blogs, reviews and ratings for analysis. However, it is difficult to
obtain sufficient and correct demographic data and daily generated content data for prediction. As a result, the prediction accuracy of anchor users is often unstable and unsatisfactory.

Recently, researchers turn to social link data to predict anchor users \cite{liu2016aligning,man2016predict}, because social link data, compared to demographic data and daily-generated content data, are much more reliable in terms of correctness and completeness. Existing works that use network structures for anchor user prediction can be divided into three categories, the unsupervised, supervised and semi-supervised models \cite{shu2017user}, according to whether or to what extent the anchor user data can be observed and collected before model training.

The unsupervised models are proposed to solve unlabeled social structure data. These methods are proved to be equivalent to the network alignment problem~\cite{singh2007pairwise,klau2009new,kollias2012network}, which falls into the NP-hard combinatorial optimization. As a consequent, these approaches are either limited to small networks or only applicable to large but sparse networks \cite{man2016predict}. The supervised models are used for predicting anchor users with training labels, which often generate accurate results when the number of labeled anchor users are adequate~\cite{kong2013inferring,man2016predict}. Last but not least, the semi-supervised models can leverage both labeled and unlabeled training data for model training. Note that in semi-supervised models the unlabeled anchor users can be predicted during the learning process~\cite{tan2014mapping,liu2016aligning}.

Obviously, the number of labeled anchor users plays a critical role in building an accurate anchor user prediction model. In order to reduce the number of training examples, Active Learning (AL) has been widely used to label training examples by iteratively selecting the most informative data for labeling at each round \cite{Settles2009Active}. In this paper, we wish to design an economic anchor user prediction model based on the active learning method. To reduce the cost of labeling user pairs in social networks, we need to address the following challenges:

\begin{itemize}
	\item \textit{Challenge 1}. In this paper, active learning algorithm builds on \textit{non-independently and identically distributed data}. Our anchor user prediction problem needs to consider social structure data obtained from multiple networks where both anchor and non-anchor user pairs exist.
	\item \textit{Challenge 2}. The anchor user pairs and the non-anchor user pairs are interlocked (or correlated). The interlock means that if $(v_A^i,v_B^m)$ is a pair of anchor users, $(v_A^i,v_B^{m'})$ and $(v_A^{i'},v_B^{m})$ are unlikely to be a pair of anchor users when $i\neq i'$ and $m\neq m'$. Hence, the query strategy of active learning should consider the interlock property among all the unlabeled user pairs.
	\item \textit{Challenge 3}. There are two components in the anchor user prediction framework, i.e., the active sampling component and the anchor user prediction component. How to iteratively reinforce the two components and minimize the labeling cost is non-trivial.
\end{itemize}

\begin{figure}[t]
	\centering
	\mbox{
		\includegraphics[width=8cm,height=9cm] {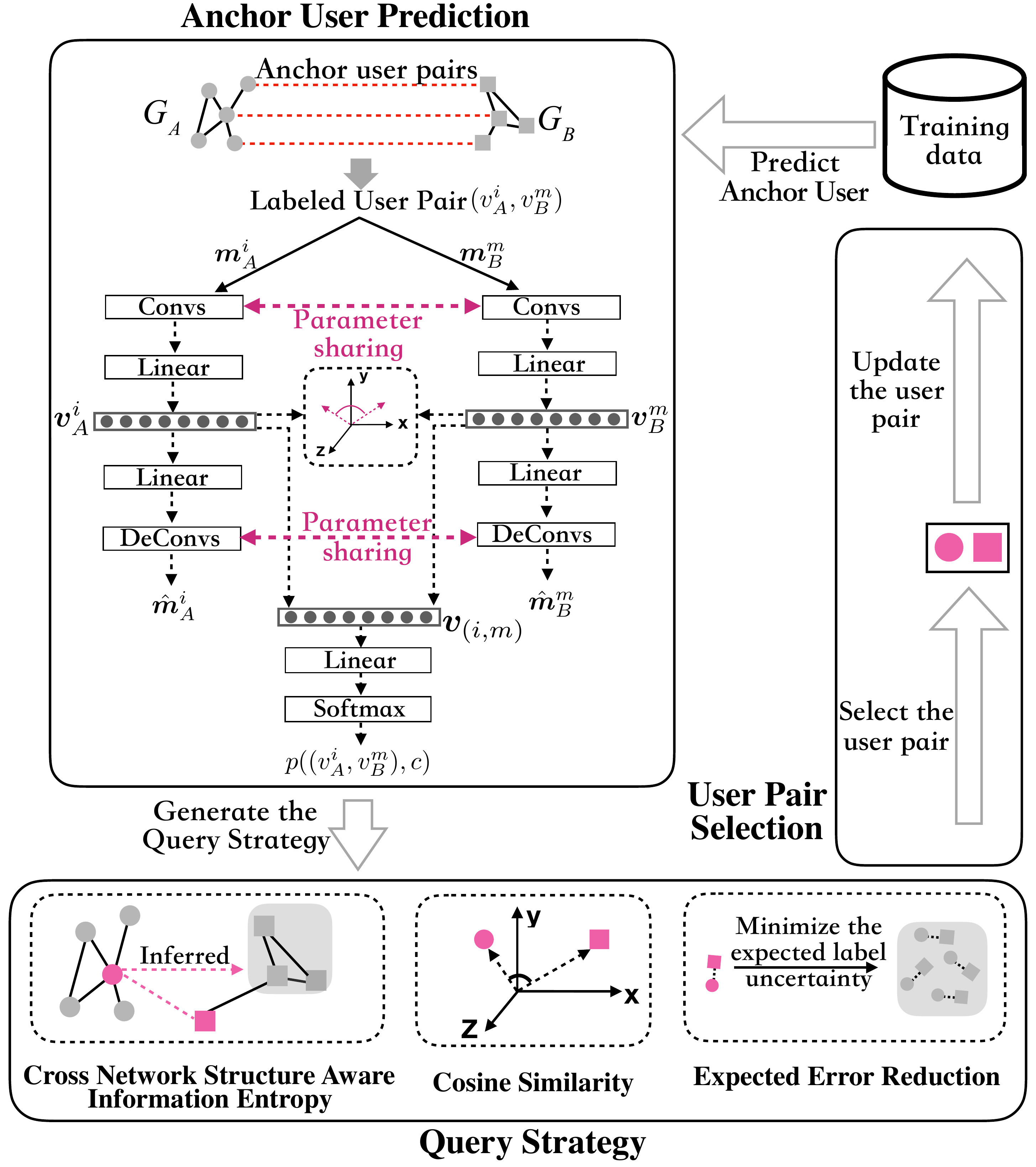}}
	\caption{An illustration of the DALAUP model
	} \vspace{-0mm}
	\label{model}
\end{figure}

To address the above challenges, we present a \underline{D}eep \underline{A}ctive \underline{L}earning model for \underline{A}nchor \underline{U}ser \underline{P}rediction (DAL\begin{small}AUP\end{small} for short). The framework of the proposed method is illustrated in Fig.~\ref{model}. From this figure, we can observe the three key components of the DAL\begin{small}AUP\end{small} method, i.e., the \textit{anchor user prediction} component, the \textit{query strategy} component and the \textit{user pair selection} component. To solve \textit{Challenge 1}, DAL\begin{small}AUP\end{small} uses a couple of convolution and deconvolution networks with shared-parameter to jointly analyze multiple social networks and obtain the representation vectors for cross-network user pairs, where structure information from both single-network and cross-network are jointly considered. To solve \textit{Challenge 2}, in the query strategy component, DAL\begin{small}AUP\end{small} uses three different criteria for active learning query, i.e., the cross network structure aware information entropy, the cosine similarity and the expected error reduction. To solve \textit{Challenge 3}, DAL\begin{small}AUP\end{small} adaptively selects user pairs by ensembling the three query methods to estimate the most informative user pairs for labeling. The main contributions of this work are summarized as follows,

\begin{itemize}
	\item We are the first to study the problem of active learning for anchor user prediction across multiple social networks, where user data are non-i.i.d. distributed.
	\item We propose a new model DAL\begin{small}AUP\end{small} by integrating active sampling and anchor user prediction. Based on the rewards of the two components, they can reinforce each other via updating parameters iteratively.
	\item We conduct experiments on real-world data sets to verify the performance of DAL\begin{small}AUP\end{small}. The results demonstrate the effectiveness of our model compared with the state-of-the-art.
\end{itemize}

\section{Preliminaries}
Given two users in two different social networks, if they share the same identity in the real-world, we call them as an anchor user pair. The anchor user prediction aims to find out these anchor user pairs.

Formally, given two social networks represented as $\mathcal{G}_A=(\mathcal{V}_A, \mathcal{E}_A)$ and $\mathcal{G}_B=(\mathcal{V}_B, \mathcal{E}_B)$, where $\mathcal{V}_A$, $\mathcal{V}_B$ are two sets of users, and $\mathcal{E}_A=\{(v_{A}^{i}, v_{A}^{j})\}$, $\mathcal{E}_B = \{(v_{B}^{m}, v_{B}^{n})\}$ are two sets of undirected edges. Denote the sizes $N_A=|\mathcal{V}_A|$ and $N_B=|\mathcal{V}_B|$. Let
$\mathcal{L} = \{(v_{A}^{i}, v_{B}^{m})\}$
denote a labeled set of pairs of anchor (or non-anchor) users, where $v_{A}^{i}\in \mathcal{G}_A$ and $v_{B}^{m}\in \mathcal{G}_B$. Let $\mathcal{U}=\{\mathcal{V}_A \otimes \mathcal{V}_B - \mathcal{L}\}$ denote an unlabeled set, where $\otimes$ is the Cartesian product.
We wish to design an active anchor user prediction model which selects a small portion of pairs of users from $\mathcal{U}$. Then, based on the labeled pairs of users, we can build accurate prediction models for anchor users.

In the active learning, we aim to design a set of query functions $\{\phi_{qs}(v_A, v_{B})\}$ and a selection function of pairs of users which can select a small portion of pairs of users $\{(v_{A}^{*}, v_{B}^{*})\}$ from $\mathcal{U}$ for labeling. This sampling and labeling process runs continuously until a given budget of labeling limit $K$ is reached. In the next section, we will formally define the learning function.

\section{The DAL\small AUP \Large Model}
We introduce the DAL\begin{small}AUP\end{small} model for anchor user prediction under the active learning setting.

\subsection{Anchor User Prediction}
Given $\mathcal{G}_{A}$, $\mathcal{G}_{B}$ and $\mathcal{L}$, we first extract the \textit{structural context} for each user in both networks $\mathcal{G}_{A}$ and $\mathcal{G}_{B}$, based on which we design a classifier for predicting the labels of pairs of users.

\subsubsection{Structural Context}

Given a social network $\mathcal{G}_{A}$, we wish to use a vector $p^{(i,s)}\in \mathbb{R}^{1\times N_{A}}$ to represent each user $v_{A}^{i}\in\mathcal{G}_{A}$, where $p^{(i,s)} $ indicates the probability of visiting $v_{A}^{j}\in\mathcal{G}_{A}$ after $s$-step transitions initiated from $v_{A}^{i}$. The transitions are usually controlled by random walks with restart \cite{tong2006fast} with a parameter $c\in (0,1)$. Specifically, each transition either returns to $v_{A}^{i}$ with probability $c$, or goes forward with probability $1-c$. As a result, $p^{(i,s)}$ can be calculated iteratively as follows,
\begin{small}
	\begin{equation}
	p^{(i,s)} = (1-c)\cdot p^{(i,s-1)}\textbf{D}_{A}+c\cdot p^{(i, 0)},
	\end{equation}
\end{small}where $\textbf{D}_{A}$ is the normalized weighted matrix of $\mathcal{G}_{A}$ with sums of the rows to be 1, and $p^{(i, 0)}$ is the initial vector with the $i$-th entry to be $1$ and the remaining entries $0$.

To capture contextual information of network structures, we define the structural context $\bm{m}_{A}^{i}$ of user $v_{A}^{i}$ to be,
\begin{small}
	\begin{equation}
	\bm{m}_{A}^{i}=\sum_{s=1}^{S}p^{(i,s)},
	\end{equation}
\end{small}where $S$ indicates the number of steps. Following the same logic, we can obtain the structural context $\bm{m}_{B}^{m}$ for each user $v_{B}^{m}\in \mathcal{V}_B$. These structural contexts are served as the input of the following neural network architecture.

\subsubsection{Representation of Pairs of Users Across Networks}
We wish to learn stable representations of pairs of users based on the structural context defined above. To preserve the structure information of a single network, we use both convolution and deconvolution neural networks to represent social networks \cite{niepert2016learning}. The convolution neural network in our model consists of $k$ convolution layers. Let $(\bm{x}^{i}_{A})^{(0)}=\bm{m}^{i}_{A}\in \mathbb{R}^{1\times N_{A}}$ be the input of user $v_{A}^{i}$ at the first convolution layer and
\begin{small}
	\begin{equation}
	(\bm{x}_{A}^{i})^{(l)}=f(\mathrm{Conv}(\textbf{W}^{(l)},(\bm{x}_{A}^{i})^{(l-1)})+b^{(l)})
	\end{equation}
\end{small}be the output of the $l$-th convolution layer, where $l=1, 2,...,k$ and $\mathrm{Conv}(\cdot)$ represents the convolution operation, $f(\cdot)$ is a non-linear activation function such as Sigmoid. $\textbf{W}^{(l)}$ is a weight parameter and $b^{(l)}$ is the bias. We use $\bm{v}_{A}^{i}=h_{\theta^{A}}((\bm{x}_{A}^{i})^{(k)})$ to denote the vector representation of user $v_{A}^{i}\in \mathcal{G}_{A}$, where $h_{\theta^{A}}(\cdot)$ is a fully connected linear layer with a parameter set $\theta^{A}=\{W_{A}, b_{A}\}$.

By using the deconvolution architecture, we can obtain the reconstruct representation $\hat{\bm{m}}_{A}^{i}$ for user $v_{A}^{i}$. Similarly, we can obtain representation $\bm{v}_{B}^{i}$ and reconstruct representation $\hat{\bm{m}}_{B}^{m}$ for user $v_{B}^{m}\in \mathcal{G}_{B}$ with the parameters $\theta^{B}=\{W_{B}, b_{B}\}$ in full connected linear layer and the shared-parameters $(\textbf{W}^{(l)},b^{(l)})_{l=1}^k$ in convolution layers.

The goal of preserving the structure information of a single network is to minimize the reconstruction error as follows,
\begin{small}
	\begin{equation}
	L_{single}=\frac{1}{\mathcal{|L|}}\sum_{(v_{A}^{i},v_{B}^{m})\in \mathcal{L}}(\|\bm{m}_{A}^{i}-\hat{\bm{m}}_{A}^{i}\|^{2}+\|\bm{m}_{B}^{m}-\hat{\bm{m}}_{B}^{m}\|^{2}).
	\end{equation}
\end{small}

We consider anchor users to be similar in their representation spaces. Then, the loss of representing cross-network information $L_{cross}$ with respect to anchor users can be formulated as follows,
\begin{small}
	\begin{eqnarray}\label{eqn:losscross}
	L_{cross}=
	& &\frac{1}{|\mathcal{L}^{+}|}\sum_{(v_{A}^{i},v_{B}^{m})\in \mathcal{L}^{+}}(1 - S(\bm{v}_{A}^{i}, \bm{v}_{B}^{m})) \\
	& &+ \frac{1}{|\mathcal{L}^{-}|}\sum_{(v_{A}^{i},v_{B}^{m})\in \mathcal{L}^{-}}max(0, S(\bm{v}_{A}^{i}, \bm{v}_{B}^{m})-\epsilon),
	\nonumber
	\end{eqnarray}
\end{small}where $S(\cdot)$ is a cosine similarity function, $\mathcal{L}^{+}\subset \mathcal{L}$ is a set of pairs of anchor users, $\mathcal{L}^{-}\subset \mathcal{L}$ is a set of pairs of non-anchor users, and $\epsilon$ is the margin.

\subsubsection{Anchor User Pair Classification}
Based on the anchor user representation $\bm{v}^{i}_{A}$ and $\bm{v}^{m}_{B}$ discussed above, we further design a classifier to estimate a pair of users $(v_{A}^{i},v_{B}^{m})$, e.g., $(v_{A}^{i},v_{B}^{m})$ is a pair of anchor users when $c=1$, otherwise, a pair of non-anchor users. The probability of a pair of users $(v_{A}^{i},v_{B}^{m})$ belongs to class $c\in\{0,1\}$ can be defined as follows,
\begin{small}
	\begin{equation}\label{equ:probability}
	p((v_{A}^{i}, v_{B}^{m}),c)=Softmax(h_{\theta}(\bm{v}_{(i,m)})),
	\end{equation}
\end{small}where $\bm{v}_{(i,m)}:=concat(\bm{v}_{A}^{i}, \bm{v}_{B}^{m})$ is an aggregated representation of a pair of users $(v_{A}^{i},v_{B}^{m})$, and $h_{\theta}(\cdot)$ is a fully connected linear layer with a parameter set $\theta=\{W,b\}$.
The loss function under the cross entropy measure can be formulated as follows,
\begin{small}
	\begin{eqnarray}
	& &L_{c}=-\frac{1}{|\mathcal{L}|}\sum_{(v_{A}^{i},v_{B}^{m})\in \mathcal{L}} \Big[y_{(v_{A}^{i},v_{B}^{m})}log(p((v_{A}^{i}, v_{B}^{m}),1))\nonumber\\
	& &~~~~~~~~~+(1-y_{(v_{A}^{i},v_{B}^{m})})log(p((v_{A}^{i}, v_{B}^{m}),0))\Big],
	\end{eqnarray}
\end{small}where $y_{(v_{A}^{i},v_{B}^{m})}\in \{0,1\}$ is the true label of a pair of anchor users $(v_{A}^{i},v_{B}^{m})$.

In other words, the objective function of anchor user prediction can be formulated by minimizing the function as follows,
\begin{small}
	\begin{equation}
	\ L_{aup}= L_{single} + \lambda_{1}L_{cross} + \lambda_{2}L_{c}+ \lambda_{3}L_{reg}
	\end{equation}
\end{small}where $L_{reg}=\frac{1}{|\mathcal{L}|}\sum_{(v_{A}^{i},v_{B}^{m})\in \mathcal{L}}(\|\bm{v}_{A}^{i}\|^{2}+\|\bm{v}_{B}^{m}\|^{2})$ is an $L_{2}$-norm regularizer term to prevent over-fitting and $\lambda_{1}$, $\lambda_{2}$, $\lambda_{3}$ are hyper-parameters to tradeoff the four parts.

\subsection{Query Strategy}
%

Based on the prediction of pairs of anchor users, we have two networks $\mathcal{G}_{A}$, $\mathcal{G}_{B}$, a labeled data set $\mathcal{L}$ and an unlabeled data set $\mathcal{U}$. Moreover, we obtain the representations of all nodes in $\mathcal{G}_{A}$, $\mathcal{G}_{B}$, and the prediction probability $p((\cdot, \cdot),c)$ for unlabeled pair in $\mathcal{U}$. Then, we wish to design query strategies to choose the most informative user pairs for labeling.

The anchor user pairs and non-anchor user pairs are interlocked. Once an anchor user pair is labeled, we can infer a large set of non-anchor user pairs. Thus, anchor user pairs are more valuable and informative for labeling than non-anchor user pairs. Here, we design two strategies \emph{cross network information entropy} and \emph{cosine similarity} to find user pairs which are more likely to be anchor ones. Moreover, we propose the third query strategy \emph{expected error reduction} to maximize the label certainty of the unlabeled use pairs.

\subsubsection{Cross Network Structure Aware Information Entropy}
Given two networks $\mathcal{G}_{A}$, $\mathcal{G}_{B}$ and a labeled set $\mathcal{L}$, according to Eq.~(\ref{equ:probability}), we can obtain the probability $p((v_{A}^{i}, v_{B}^{m}), c)$ of the candidate user pair $(v_{A}^{i}, v_{B}^{m})\in \mathcal{U}$ belonging to class $c$.
The measure function for unknown label $c$ of each unlabeled user pair $(v_{A}^{i}, v_{B}^{m})$ is defined as follows,
\begin{small}
	\begin{equation}
	\phi_{saie}(v_{A}^{i}, v_{B}^{m}) =
	\sum_{c\in\{0,1\}}\Big[p((v_{A}^{i}, v_{B}^{m}),c) \cdot R((v_{A}^{i}, v_{B}^{m}),c)\Big],
	\label{saie}
	\end{equation}
\end{small}where $R((v_{A}^{i}, v_{B}^{m}),c)$ is the reward function to measure the labeling outcome $((v_{A}^{i}, v_{B}^{m}),c)$. For information entropy, reward function $R((v_{A}^{i}, v_{B}^{m}),c)$ can be defined as follows,
\begin{small}
	\begin{equation}\label{equ:reward}
	R((v_{A}^{i}, v_{B}^{m}),c)=-\log p((v_{A}^{i}, v_{B}^{m}),c).
	\end{equation}
\end{small}

Due to the interlock between anchor user pairs and non-anchor user pairs, the reward for labeling anchor user pairs is much more important than labeling non-anchor ones. Let $\mathcal{S}_{(v_{A}^{i}, v_{B}^{m})}$ denote a set of non-anchor user pairs inferred from $(v_{A}^{i}, v_{B}^{m})$ when it is labeled to be anchor one, we modify the reward function in Eq.~(\ref{equ:reward}) to be,
\begin{small}
	\begin{equation*}
	R((v_{A}^{i}, v_{B}^{m}),0) = -\log p((v_{A}^{i}, v_{B}^{m}),0)
	\end{equation*}
	and
	\begin{equation}\label{saieR}
	R((v_{A}^{i}, v_{B}^{m}),1) =
	-\log p((v_{A}^{i}, v_{B}^{m}),1)-\sum_{s\in \mathcal{S}_{(v_{A}^{i}, v_{B}^{m})}}\log p(s,0).
	\end{equation}
\end{small}

Based on the new measure functions, we enable the cross network information entropy for candidate user pairs in $\mathcal{U}$, i.e., the larger $\phi_{ie}(v_{A}^{i}, v_{B}^{m})$, the more informative $(v_{A}^{i}, v_{B}^{m})$ we obtain.

\subsubsection{Cosine Similarity}
After predicting the anchor users, we obtain the representations of $v_{A}^{i}$, $v_{B}^{m}$ in the low-dimensional vector space. Since the anchor user pairs are similar between their representation vectors as in Eq.~(\ref{eqn:losscross}), we adopt the cosine similarity $cs(v_{A}^{i}, v_{B}^{m})$ to query each candidate user pair $(v_{A}^{i}, v_{B}^{m})\in \mathcal{U}$.
The cosine similarity based query strategy $\bm{\phi}_{cs}$ is defined as follows,
\begin{small}
	\begin{equation}
	\phi_{cs}(v_{A}^{i}, v_{B}^{m}) =|cs(\bm{v}_{A}^{i}, \bm{v}_{B}^{m})|.
	\label{cs}
	\end{equation}
\end{small}The larger $\phi_{cs}(v_{A}^{i}, v_{B}^{m})$ is, the more valuable $(v_{A}^{i}, v_{B}^{m})$ is.

\subsubsection{Expected Error Reduction}
For $(v_{A}^{i}, v_{B}^{m})\in \mathcal{U}$, we introduce the expected error reduction query strategy, based on the prediction performance on the remaining unlabeled instances $\mathcal{U}\backslash{(v_{A}^{i}, v_{B}^{m})}$ \cite{aggarwal2014active}. The expected error reduction aims to choose $(v_{A}^{i}, v_{B}^{m})\in \mathcal{U}$ to maximize the label certainty of use pairs in $\mathcal{U}\backslash{(v_{A}^{i}, v_{B}^{m})}$. Intuitively, the estimated label of $(\hat{v}_{A}^{i}, \hat{v}_{B}^{m})\in\mathcal{U}\backslash{(v_{A}^{i}, v_{B}^{m})}$ relates the probability $\hat{p}((\hat{v}_{A}^{i}, \hat{v}_{B}^{m}),c)$. Thus, the error reduction based query for $(v_{A}^{i}, v_{B}^{m})$ can be defined as follows,

\begin{small}
	\begin{eqnarray}\label{eer}
	& &\phi_{eer}(v_{A}^{i}, v_{B}^{m})=\sum_{c\in\{0,1\}}p((v_{A}^{i}, v_{B}^{m}),c)\cdot\\
	& &~~~~~~~~~~~~\Big(\sum_{c\in\{0,1\}}\sum_{(\hat{v}_{A}^{i}, \hat{v}_{B}^{m})\in\mathcal{U}\backslash{(v_{A}^{i}, v_{B}^{m})}}|\hat{p}((\hat{v}_{A}^{i}, \hat{v}_{B}^{m}),c)-0.5|\Big) , \nonumber
	\end{eqnarray}
\end{small}where $\hat{p}((\hat{v}_{A}^{i}, \hat{v}_{B}^{m}),c)$ is the probability that user pair $(\hat{v}_{A}^{i}, \hat{v}_{B}^{m})$ belongs to class $c$, which can be calculated by Eq.~(\ref{equ:probability}) with $((v_{A}^{i}, v_{B}^{m}),c)$ being added to $\mathcal{L}$. The bigger $\phi_{eer}(v_{A}^{i}, v_{B}^{m})$ is,
the more helpful is $(v_{A}^{i}, v_{B}^{m})$ for maximizing the label certainty of pairs in $\mathcal{U}\backslash{(v_{A}^{i}, v_{B}^{m})}$.

\subsection{Active User Pair Selection Mechanism}
Based on the above query strategies, a natural question is how to choose one candidate at each iteration $e$ for labeling. As the prediction performance tends to get better when the labeled data set $\mathcal{L}$ becomes larger after adding the labeled user pairs at each iteration, it is intuitive that the rewards are not independent random variables at different iterations. In this paper the reward $R_e(\phi)$ is defined as the average rise of Precision@30 and MAP@30, which can be estimated on the validate data. The recommended user pairs of the strategy are influenced by the recommendation results at previous iterations.

To simulate the adversarial setting, we use the multi-armed bandit method as the solution \cite{auer2002nonstochastic}. To adaptively select the most informative user pairs, we adjust the $\epsilon$-greedy algorithm to make a trade-off between exploitation and exploration, where the exploration probability $\epsilon$ is time-sensitive, e.g. $\epsilon \sim Beta(0.1, e)$ for different iteration $e\in \{1,2,\ldots,K/bs\}$. At the $e$-th iteration, we first sample an $\epsilon$ according to $Beta(0.1, e)$. The selection randomly selects a query strategy with probability $\epsilon$ (Exploration); Otherwise, it selects the one with the highest historical mean reward (Exploitation).

Specifically, let $Q_{e-1}(\phi)$ denote the historical mean reward of the query strategy $\phi$ at the first $e-1$ active learning iterations, then the query strategy $\phi_{e}$ at the $e$-th iteration can be chosen as below,
\begin{small}
	\begin{equation}
	\phi_{e}=\left\{
	\begin{array}{lcl}
	rand(\phi_{saie}, \phi_{cs}, \phi_{eer}) & & {\mbox{with prob.}~ \epsilon}\\
	\mathop{\arg\max}_{\phi \in \{\phi_{saie}, \phi_{cs}, \phi_{eer}\}}Q_{e-1}(\phi) & & {\mbox{otherwise}},
	\end{array} \right.
	\end{equation}
\end{small}where $\epsilon$ is a sample from $Beta(0.1, e)$.
With the iteration time $e$ increasing, $\epsilon$ is expected to decrease, which will reduce the attention onto the exploitation.

Let $n_{e-1}(\phi)$ be the time of query strategy $\phi$ employed at the first $e-1$ iteration.
Assume the query strategy $\phi_{saie}$ is employed in the $e$-th iteration, then the historical mean reward and employ times of $\phi_{saie}$ can be updated through,
\begin{small}
	\begin{equation}
	Q_{e}(\phi) := \frac{n_{e-1}(\phi)\cdot Q_{e-1}(\phi)+R_e(\phi)}{n_{e-1}(\phi)+1}
	\label{Q}
	\end{equation}
\end{small}and $n_{e}(\phi):=n_{e-1}(\phi)+1$ with $\phi=\phi_{saie}$. Meanwhile, $Q_{e}(\phi_{cs}):=Q_{e-1}(\phi_{cs})$, $Q_{e}(\phi_{eer}):=Q_{e-1}(\phi_{eer})$, $n_{e}(\phi_{cs}):=n_{e-1}(\phi_{cs})$ and $n_{e}(\phi_{eer}):=n_{e-1}(\phi_{eer})$. The update is similar, when $\phi_{cs}$ or $\phi_{eer}$ is employed at the $e$-th iteration.

\subsection{Algorithms}
The algorithm for solving DAL\begin{small}AUP\end{small} is summarized in Algorithm~\ref{dalaup}, where $\bm{\Theta}^{e}$ denotes a set of the learnable parameters in the \emph{anchor user prediction} component at the $e$-th iteration, where $e\in\{0,1,\ldots,K/bs\}$.

\begin{algorithm}[htb]
	\caption{The DAL\small AUP \normalsize algorithm}
	\label{dalaup}
	\begin{algorithmic}[1]
		\REQUIRE ~~\\ Network $\mathcal{G}_{A}$ and $\mathcal{G}_{B}$, labeled and unlabeled set $\{\mathcal{L}, \mathcal{U}\}$, maximum size $K$, size of small batch set $bs$, and query strategy set $\Phi:=\{\phi_{saie}, \phi_{cs}, \phi_{eer}\}$.
		\ENSURE ~~\\
		Parameter set $\bm{\Theta}^{K/bs}$ and ultimate prediction results.\\
		\STATE {Extract structural context $\bm{m}_{A}, \bm{m}_{B}$}.
		\STATE {Learn initial parameter $\bm{\Theta}^{0}$ with initial $\mathcal{L}$}.
		\STATE {Initialize $n_{0}(\phi)\leftarrow 0, Q_{0}(\phi)=0$ for each $\phi\in\Phi$}.
		\STATE {Initialize $e\leftarrow1$}.
		\WHILE {$e \le K/bs$}
		\STATE{$\epsilon\sim Beta(0.1,e)$, $\gamma\sim U[0,1] $}
		\IF{$\gamma< \epsilon$}
		\STATE{$\phi_e \leftarrow rand(\Phi)$}
		\ELSE
		\STATE {$\phi_e \leftarrow \mathop{\arg\max}_{\phi'\in\Phi}Q_{e-1}(\phi')$}
		\ENDIF
		\STATE {Estimate $R_{e}(\phi_{e})$ on validate data set}.
		\STATE {Update $Q_{e}(\phi_e)$ with Eq.~(\ref{Q})}.
		\STATE {$n_{e}(\phi_{e})\leftarrow n_{e-1}(\phi_{e})+1$}.
		\STATE {$n_{e}(\phi^{*})\leftarrow n_{e-1}(\phi^{*})$ for each $\phi^{*}\in\Phi\backslash\phi_{e}$}.
		\STATE {$Q_{e}(\phi^{*})\leftarrow Q_{e-1}(\phi^{*})$ for each $\phi^{*}\in\Phi\backslash\phi_{e}$}.
		\STATE {Choose $bs$ candidate user pairs $\mathcal{C}\subseteq\mathcal{U}$ according to query strategy $\phi_{e}$ and label them}.
		\STATE {$\mathcal{L}\leftarrow\mathcal{L}\cup\mathcal{C}, \mathcal{U}\leftarrow\mathcal{U}\backslash\mathcal{C}$}.
		\STATE {Learn parameter $\bm{\Theta}^{e}$ with updated $\mathcal{L}$}.
		\STATE {$e\leftarrow e+1$}.
		\ENDWHILE
		\RETURN{$\bm{\Theta}^{K/bs}$ and ultimate prediction results}.
	\end{algorithmic}
\end{algorithm}

\section{Experiments}
In this section, we compare DAL\begin{small}AUP\end{small} with existing baseline methods on real world social networks. Experimental results show the effectiveness of DAL\begin{small}AUP\end{small} compared with the state-of-the-art methods.
\footnote{https://github.com/chengaf/DALAUP}.

\subsection{Data Sets}
We use Foursquare and Twitter \cite{kong2013inferring} as the testbed. All the anchor user pairs are known in these data sets. The statistics are listed in Table 1.

We consider the labeled anchor user pairs as positive data and the inferred or labeled non-anchor user pairs as negative data. In the training process, we adopt undersampling to balance positive and negative data \cite{japkowicz2002class}. In our experiment, we randomly sample two non-anchor user pairs corresponding to an anchor user pair.

\begin{table}[htbp]
	\setlength{\abovecaptionskip}{0.2cm}
	\begin{tabular}{cccc}
		\toprule
		Network&User&Relation&Anchor User \\
		\midrule
		Foursquare&5,313&76,972&\multirow{2}{*}{3,141} \\
		Twitter&5,120&164,919& \\
		\bottomrule
	\end{tabular}
	\centering \caption{Statistics of Foursquare and Twitter\label{dataset}}
\end{table}

\subsection{Baseline Methods and Evaluation Metrics}
In order to evaluate the effectiveness of DAL\begin{small}AUP\end{small}, we choose the state-of-the-art unsupervised, supervised and semi-supervised methods for comparison,
\begin{itemize}
	\item \textbf{DW:} DeepWalk \cite{perozzi2014deepwalk}. An unsupervised method which first represents two social networks to a low-dimensional vector space and then calculates cosine similarity between vector representations of cross-network user pairs to predict anchor user.
	\item \textbf{PALE:} Predicting Anchor Links via Embedding \cite{man2016predict}. A supervised model that learns network embedding with extended structural information.
	\item \textbf{IONE:} Input-output Network Embedding \cite{liu2016aligning}. A semi-supervised approach that gets aligned vector representations for multiple networks.
\end{itemize}

In addition, we introduce some variants of DAL\begin{small}AUP\end{small} for comparison, including
\begin{itemize}
	\item \textbf{AUP:} Anchor User Prediction method proposed in this paper, which is a supervised manner.
	\item \textbf{DAL\begin{small}AUP\end{small}-saie, DAL\begin{small}AUP\end{small}-eer, DAL\begin{small}AUP\end{small}-cs, DAL\begin{small}AUP\end{small}-ie, and DAL\begin{small}AUP\end{small}-Rand.:} the variants of DAL\begin{small}AUP\end{small} distinguished by their different AL query strategies, i.e., DAL\begin{small}AUP\end{small}-saie ia an active anchor user prediction model with only one query strategy $\phi_{saie}$ defined by Eq.~(\ref{saie}) and Eq.~(\ref{saieR}). Same argument for DAL\begin{small}AUP\end{small}-eer and DAL\begin{small}AUP\end{small}-cs. Besides, DAL\begin{small}AUP\end{small}-ie is a simplified model of DAL\begin{small}AUP\end{small}-saie with a query strategy $\phi_{ie}:=\phi_{saie}$ in which $R$ is defined by Eq.~(\ref{equ:reward}). The last method DAL\begin{small}AUP\end{small}-Rand. is to randomly select user pairs from unlabeled set to label.
\end{itemize}

For each user in Foursquare, all above anchor user prediction algorithms output a list of candidate anchor users in Twitter.
The Precision@30 \cite{liu2016aligning} and MAP@30 \cite{man2016predict} are used as metrics for performance comparison.

\subsection{Parameter Setting}
The optimal parameter settings for each method are either determined by experiments or taken from the suggestions by previous works. Following \cite{perozzi2014deepwalk}, we use the default parameter setting for DW, i.e., window size is $5$ and walks per user is $80$. In our method, we set the restart probability to $c=0.6$ \cite{tong2006fast} and the number of convolution layers $k=2$. 
Margin $\epsilon$ in $\mathcal{L}_{cross}$ is set to $0$.
The other parameters are set as: $\lambda_{1}=0.01$, $\lambda_{2}=0.01,\ \lambda_{3}=10^{-5}$.

\subsection{Results and Analysis}
To evaluate the effectiveness of DAL\begin{small}AUP\end{small}, we conduct two parts of experiments: effectiveness of AUP and effectiveness of DAL\begin{small}AUP\end{small}. For all experiments, we repeat the process $10$ times and report the average results to test the statistical significance of the comparison results.

\subsubsection{Effectiveness of AUP}
\begin{figure}[t]
		\setlength{\abovecaptionskip}{0.1cm}
		\setlength{\belowcaptionskip}{0cm}
		\centering
	\mbox{
		\includegraphics[width=7.5cm] {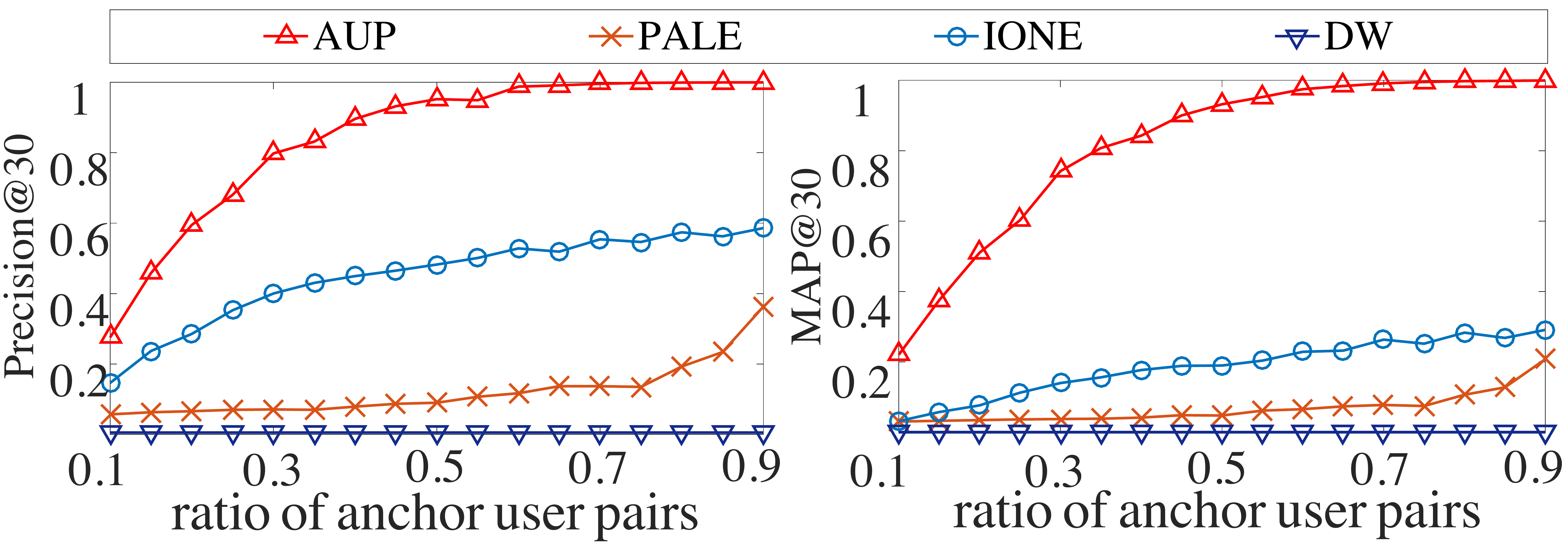}}
	\caption{The performance comparisons of different anchor link prediction methods}
\label{ratio}
\end{figure}

In this experiment, different ratios of anchor user pairs are sampled randomly as positive data and the corresponding inferred non-anchor user as negative data for training. The remaining ones are viewed as test data. We set ratio $\eta \in\{0.1,0.15,0.2,...,0.9\}$ respectively and fix the representation dimension $d=56$ \cite{liu2016aligning}. Experiment results are shown in Fig.~\ref{ratio}.

From Fig.~\ref{ratio}, we can observe that the performance of our supervised method AUP is significantly better than the other comparing methods. Specially, when the ratio $\gamma$ comes to $0.5$, IONE can achieve about $48\%$ and $19\%$ in Precision@30 and MAP@30, while AUP can obtain $97\%$ and $95\%$ in these two measures.

\subsubsection{Effectiveness of DAL\small AUP \normalsize}
In this part, we investigate the effectiveness of the AL query strategies and the select mechanism proposed in section 3.2 and 3.3. To ensure that the performance difference in the experiments is solely caused by
different AL query strategies and their select mechanism,
we first randomly split the whole anchor user data into four parts: an initial training set to build an initial model, a validate set to get the rewards of different query strategies, a test set to evaluate the performance of the model, and an unlabeled set to select user pairs \cite{shen2004multi}.

\begin{figure}[t]
	\setlength{\abovecaptionskip}{0.1cm}
	\setlength{\belowcaptionskip}{0.cm}
	\centering
	\mbox{
		\includegraphics[width=8cm] {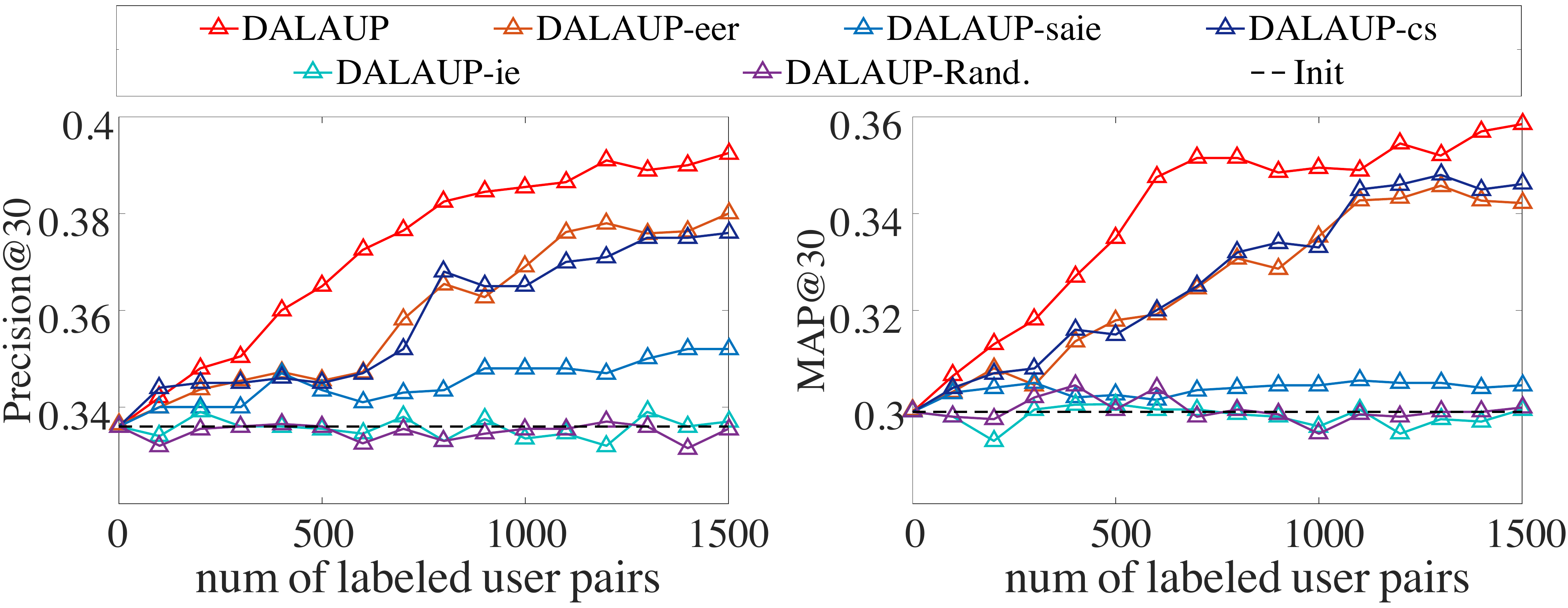}}
	\caption{The comparisons of different query strategies}
	\label{al}
\end{figure}

The size of anchor user pairs in each part is 100, 300, 600 and 2,141 respectively. According to the interlock property, the non-anchor user pairs can be inferred from the anchor user pair set. At each iteration, we select a small batch of user pairs by following our AL query strategies, add the real label on them, and put them into the training set. The batch size $bs=100$ and the maximum size $K=1,500$.
The performance comparisons of different query strategies with the different number of queried user pairs
are shown in Fig.~\ref{al}, in which the dotted line is the performance of the AUP model without any newly labeled information being added.

From Fig.~\ref{al}, we can observe that:
\begin{itemize}
	\item DAL\begin{small}AUP\end{small} significantly outperforms all the other methods in terms of both Precision@30 and MAP@30, which demonstrates that the proposed active user pair selection methods is effective in improving the prediction performance.
	
	\item The poor results of DAL\begin{small}AUP\end{small}-Rand. further validate the effectiveness of the proposed AL query strategies.
	
	\item Compared to information entropy DAL\begin{small}AUP\end{small}-ie, the modified DAL\begin{small}AUP\end{small}-saie method largely improves
	Precision@30 and MAP@30
	, which proves that cross network structure information can be used to ameliorate information entropy.
\end{itemize}

In addition, we investigate the performance of different anchor link prediction methods when adding 1500 extra labeled user pairs. These labeled data are selected in two manners. One is to select by DAL\begin{small}AUP\end{small} , while the other is to select randomly. For the convenience comparison, we also add the performance of these methods with no extra labeled data added.

\begin{figure}[t]
	\setlength{\abovecaptionskip}{0.1cm}
	\setlength{\belowcaptionskip}{0.cm}
	\centering
	\mbox{
		\includegraphics[width=8cm] {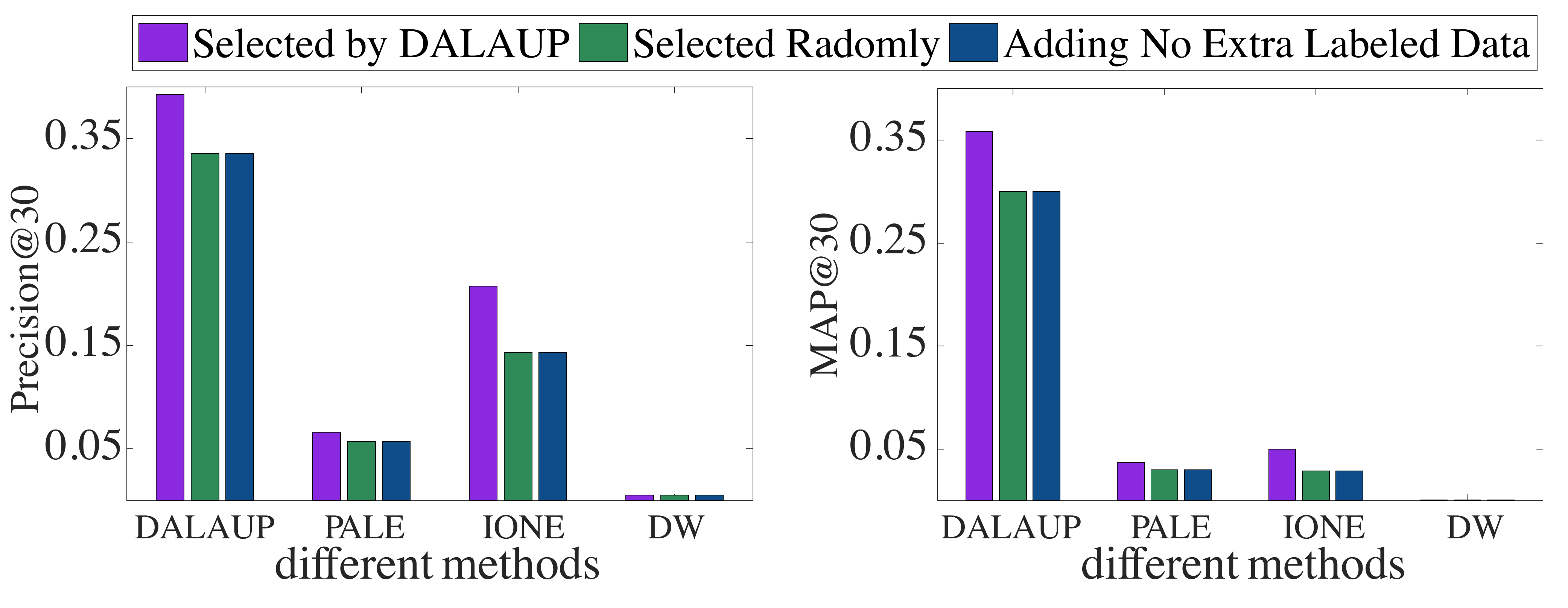}}
	\caption{The comparisons of anchor link prediction methods when labeling 1500 user pairs}
	\label{qsrand}
\end{figure}

The results are shown in Fig.~\ref{qsrand}, from which we can observe that:
\begin{itemize}
	\item DAL\begin{small}AUP\end{small} outperforms all the other methods, no matter the data selected by itself or by random.
	\item The labeled data, selected by DAL\begin{small}AUP\end{small}, can improve performance for the baselines.
	\item The randomly selected 1500 data for labeling cannot help better the benchmark methods, which confirms the importance of active selection methods.
\end{itemize}


\section{Conclusion}
In this paper, we present a deep active learning based method DAL\begin{small}AUP\end{small} to solve the anchor user prediction problem. We design a couple of convolution and deconvolution networks with shared-parameter to estimate the vector representations of cross-network anchor users. To solve the correlation difficulty of labeling anchor user pairs and non-anchor user pairs, we design three query strategies and a time-sensitive user pair selection algorithm for precisely selecting pairs of users for labeling . Experiments on real-world social network datasets demonstrate the effectiveness and efficiency of the proposed DAL\begin{small}AUP\end{small} method.

\section*{Acknowledgments}
This work was supported in part by the National Key Research and Development Program of China (No. 2016YFB0801301), the NSFC (No. 61872360), 
the MQNS (No. 9201701203), the MQEPS (No. 90275252 and No. 96804590), the MQRSG (No. 95109718), the Youth Innovation Promotion Association CAS (No. 2017210), and the Investigative Analytics Collaborative Research Project between Macquarie University and Data61 CSIRO.

\bibliography{ijcai19}
\bibliographystyle{ijcai19}

\end{document}